\documentclass[10pt,twocolumn,letterpaper]{article}

\usepackage[pagenumbers]{iccv} % To force page numbers, e.g. for an arXiv version

\definecolor{iccvblue}{rgb}{0.21,0.49,0.74}
\usepackage[pagebackref,breaklinks,colorlinks,allcolors=iccvblue]{hyperref}

\usepackage{float}
\usepackage{cuted}
\usepackage{multirow}
\usepackage{mathtools}
\usepackage{wrapfig}
\newcommand{\argtopk}{\mathop{\mathrm{arg\,top}\,k}}
\def\paperID{14264} % *** Enter the Paper ID here
\def\confName{ICCV}
\def\confYear{2025}

\title{{MAPEX}: Modality-Aware Pruning of Experts for \\ Remote Sensing Foundation Models}

\author{
Joëlle Hanna \thanks{Both authors contributed equally to this work} \quad Linus Scheibenreif \footnotemark[1] \quad Damian Borth \\
AIML Lab, School of Computer Science
\\
University of St.Gallen \\
{\tt\small \{joelle.hanna, linus.scheibenreif, damian.borth\}@unisg.ch}
}

\begin{document}
\maketitle
\begin{abstract}

Remote sensing data is commonly used for tasks such as flood mapping, wildfire detection, or land-use studies. For each task, scientists carefully choose appropriate modalities or leverage data from purpose-built instruments. Recent work on remote sensing foundation models pre-trains computer vision models on large amounts of remote sensing data. These large-scale models tend to focus on specific modalities, often optical RGB or multispectral data. For many important applications, this introduces a mismatch between the application modalities and the pre-training data. Moreover, the large size of foundation models makes them expensive and difficult to fine-tune on typically small datasets for each task. We address this mismatch with {MAPEX}, a remote sensing foundation model based on mixture-of-modality experts. {MAPEX} is pre-trained on multi-modal remote sensing data with a novel modality-conditioned token routing mechanism that elicits modality-specific experts. To apply the model on a specific task, we propose a modality aware pruning technique, which only retains experts specialized for the task modalities.
This yields efficient modality-specific models while simplifying fine-tuning and deployment for the modalities of interest. We experimentally validate {MAPEX} on diverse remote sensing datasets and show strong performance compared to fully supervised training and state-of-the-art remote sensing foundation models.
Code is available at \href{https://github.com/HSG-AIML/MAPEX}{github.com/HSG-AIML/MAPEX}.

\end{abstract}    
\section{Introduction}
\label{sec:intro}

% \begin{itemize}
%     \item remote sensing foundation models are trained for single modalities, mostly RGB.
%     \item existing foundation models are big and hard to handle for downstream tasks (need lots of memory, compute, FT data, ...)
%     \item often, we'd like to use foundation models for very specific downstream tasks that are solved with very specific modalities. NIR in wildfire, S1 in floods, RGB in land-cover, SWIR in agriculture, ...
%     \item hence, there's a mismatch between existing pre-trained models are their real-world use cases
%     \item we address this in two-steps: 1) pre-training method for mixture-of-experts model with multi-modal RS data 2) pruning technique to extract modality-specific expert models
%     \item Therefore, we combine best of both worlds: leverage large multi-modal data for pre-training to the benefit of all downstream modalities but also provide small, specialized models that are easy to use and specifically pre-trained for a wide variety of modalities.
% \end{itemize}

\begin{figure}[t]
    \centering
    \includegraphics[width=0.9\linewidth]{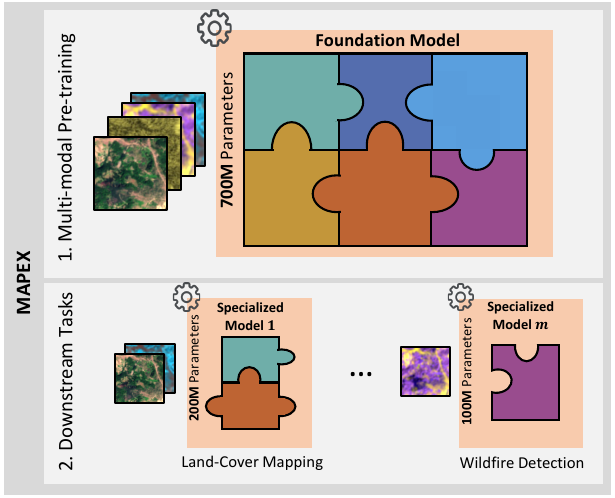}
\caption{Overview of MAPEX. A multi-modal foundation model (top) is pretrained using a Mixture-of-Experts architecture with modality-aware routing. During fine-tuning (bottom), irrelevant experts are pruned, yielding smaller, specialized models tailored to specific downstream tasks (e.g., land-cover mapping, wildfire detection, etc.), making them lighter and easier to fine-tune.}
    \label{fig:mixture-module}
\end{figure}

Computer vision methods are widely applied in the remote sensing domain for tasks such as classification, segmentation, or detection in imagery obtained from satellites or aerial platforms~\cite{zhu2017deep}. Increasingly, visual foundation models tailored to the characteristics of remote sensing imagery, such as top-down viewpoint, varying resolution, or multi-spectral data, are adopted for these tasks~\cite{xiao2024foundation,lu2024ai}. Advanced foundation models pre-trained on unlabeled data with self-supervised objectives present a great opportunity in the remote sensing domain, where labeled datasets are small and expensive, but unlabeled data is widely available in huge quantities~\cite{chi2016big}. This is particularly important for tasks such as land cover mapping, where pixel-level labels are costly to obtain and weak supervision remains the norm ~\cite{Hanna2023SparseMV}.
Existing foundation models for remote sensing leverage this data to provide strong pre-trained models for downstream tasks but remain limited in their use of multi-modal remote sensing data. Most remote sensing foundation models are pre-trained on individual modalities such as optical RGB or multi-spectral imagery, which limits their use for tasks on the many other remote sensing modalities~\cite{Cong2022SatMAEPT,Reed2022ScaleMAEAS}. On the other hand, foundation models that are designed to leverage multi-modal data for pre-training tend to be large, expensive to train, or difficult to use for downstream applications where the whole suite of modalities used for pre-training is typically not available~\cite{nedungadi2024mmearth,yan2023ringmo,guo2024skysense}.

\begin{figure*}[t]
    \centering
    \includegraphics[width=0.9\linewidth]{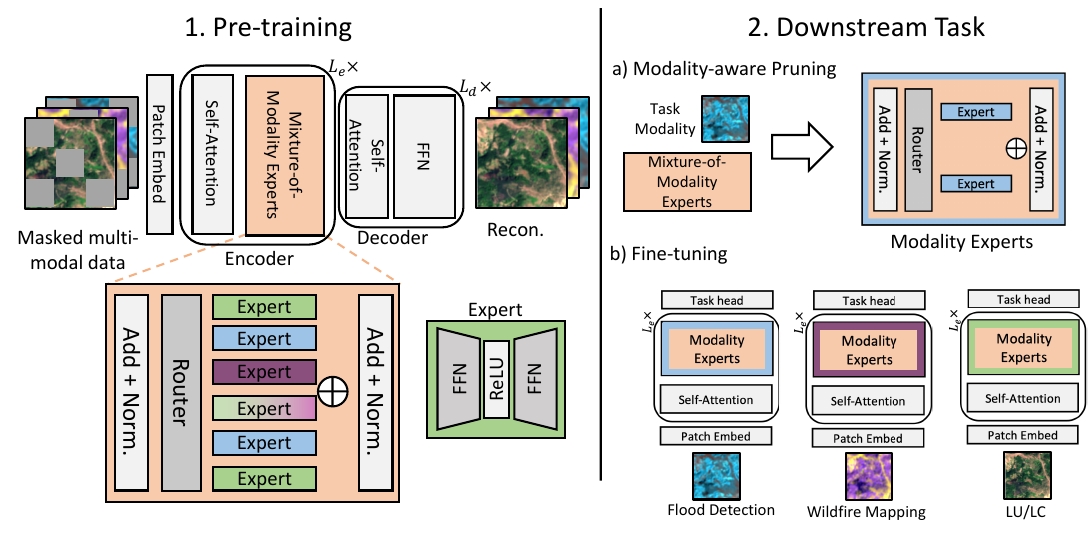}
    \caption{Overview of the pre-training and downstream application process. \textbf{1.} The model is trained with all modalities on a masked reconstruction task. Modalities are processed by specialized experts in the mixture-of-modality experts module. \textbf{2.a)} Modality experts are pruned for a modality of interest. \textbf{2.b)} The resulting modality experts model is fine-tuned on a downstream task of the corresponding modality.}
    \label{fig:overview}
    \vspace{-1em}
\end{figure*}

Models that exclusively work with multiple modalities can be restrictive in remote sensing, where instruments and satellite missions are often designed to solve specific tasks, such that multi-modal data might not be necessary downstream. For instance, synthetic aperture radar (SAR) data is commonly used to map flooding because it is sensitive to water cover and able to penetrate clouds, which are common around flood events~\cite{amitrano2024flood}. Pairing SAR with optical data that cannot penetrate cloud cover for the sake of maintaining the multi-modal pre-training setup for downstream fine-tuning might not be useful in such cases. Similar arguments could be made for a series of other remote sensing tasks in areas ranging from wildfire detection~\cite{cocke2005comparison} to environmental monitoring or agricultural applications~\cite{sishodia2020applications}.
Multi-modal pre-training gives computer vision models a more complete picture of the world and allows them to leverage valuable synergies between instruments~\cite{baltruvsaitis2018multimodal}. However, there remains a mismatch between the pre-training stage and the real-world use cases of big remote sensing foundation models.
We \textbf{address} this mismatch in two steps (see Figure~\ref{fig:overview}): first, we propose a novel pre-training method for multi-modal remote sensing data based on a mixture-of-experts architecture with modality-conditioned token routing. This method facilitates the use of multi-modal pre-training data while enforcing modality-specific structure in large parts of the pre-trained model.
% In the second step, we leverage this modality-specific structure to prune parts of the model that are not relevant for the modality of the downstream task.
In the second step, we leverage this modality-specific structure with \textbf{M}odality \textbf{A}ware \textbf{P}runing of \textbf{E}xperts (\textbf{MAPEX}) for the modality of the downstream task (see Figure~\ref{fig:mixture-module}).
This yields a significantly smaller, more focused model that is specialized for one (or multiple) modalities of interest and can be easily fine-tuned on that modality.
This approach allows us to combine the best of both worlds: leverage large-scale multi-modal data for self-supervised pre-training of large foundation models, but also provide small, specialized models that are easy to use and pre-trained for a wide variety of modalities.
In summary, the \textbf{contributions} of our work are as follows:
\begin{itemize}
    \item New multi-modal pre-training approach for a remote sensing mixture-of-experts model.
    \item Novel modality-conditioned expert routing method to create modality specific experts within the multi-modal model.
    \item Pruning technique to generate flexible modality-specific models tailored to different downstream tasks.
\end{itemize}

\section{Related Work}
\label{sec:related_work}

\paragraph{Foundation Models}
% In general, vision, language. How are multiple modalities handled. Pros and cons. 
Foundation models are based on the idea of training big models on large, unlabeled data with self-supervised learning techniques before adapting them to specific downstream tasks~\citep{bommasani2021opportunities}. This approach has produced strong, general models for natural language~\citep{brown2020language}, computer vision~\citep{kirillov2023segment}, and more specialized variants for domains such as remote sensing~\citep{hong2024spectralgpt}. Increasingly, research in foundation models has focused on multi-modal data, particularly the combination of language and visual data~\citep{radford2021learning}. Multi-modal foundation models benefit from training on datasets of multiple paired modalities, producing models that perform well on individual modalities and enabling novel applications with multi-modal data (see ~\cite{li2024multimodal} for a review). Large scale pre-training improves performance on downstream applications, but also requires big models for optimally compute-efficient training~\cite{kaplan2020scaling}. This increases latency as well as deployment and model serving costs, limiting foundation model use in time or cost sensitive settings. A growing body of literature addresses foundation model efficiency issues with techniques like pruning~\cite{sietsma1988neural}, distillation~\cite{hinton2015distilling} or parameter-efficient fine-tuning methods~\cite{hu2021lora}.

\paragraph{Mixture-of-Experts}
% What's the main idea, how does it generally work, where is it successful.
% A complementary area of research focuses on conditional computation as a way of increasing model size efficiently. Mixture-of-experts (MoE) models are commonly used in large foundation models~\cite{shazeer2017outrageously}. These models consist of multiple \textit{expert} modules and a \textit{router} that assigns tokens to each expert during training and inference. In standard MoE models, the router $r$ produces logits from tokens $t$ as $r(t)=W_r\cdot t$, which are turned into routing probabilities with a softmax to determine the routing of $t$. In transformer models, MoEs replace feed-forward network layers (FFN) with multiple experts, each implemented as a FFN itself.

% This approach has been used to train performant large-scale language~\cite{fedus2022switch} and vision models~\cite{riquelme2021scaling}, but recently also received attention in the remote sensing domain~\cite{hanmultisensory, cao2024mv, lin2024rs}. Sparse mixture-of-expert models conditionally route tokens to a subset of expert modules, making inference more efficient~\cite{fedus2022switch}. Similarly, large-scale distributed training can benefit from conditional computation with distinct model components facilitating parallelism across GPUs~\cite{shazeer2017outrageously}. In the multi-modal context, mixture-of-experts models have been used in language-image models~\cite{lin2024moma}, incorporating the idea of modality-specific experts with token routing based on each token’s modality.
A complementary area of research focuses on conditional computation as a way of increasing model size efficiently. Mixture-of-experts (MoE) models are commonly used in large foundation models~\cite{shazeer2017outrageously}. These models consist of multiple \textit{expert} modules and a \textit{router} that assigns tokens to each expert during training and inference. In standard MoE models, the router $r$ produces logits from tokens $t$ as $r(t)=W_r\cdot t$, which are turned into routing probabilities with a softmax to determine the routing of $t$. In transformer models, MoEs replace feed-forward network layers (FFN) with multiple experts, each implemented as a FFN itself. This approach has been used to train performant large-scale language~\cite{fedus2022switch} and vision models~\cite{riquelme2021scaling} and has recently gained traction in remote sensing~\cite{hanmultisensory, cao2024mv, lin2024rs}. Sparse MoE models conditionally route tokens to a subset of expert modules, making inference more efficient~\cite{fedus2022switch}, while large-scale distributed training benefits from conditional computation by facilitating parallelism across GPUs~\cite{shazeer2017outrageously}. 

In the multi-modal context, MoEs have been applied to language-image models~\cite{lin2024moma}, incorporating modality-specific experts with token routing based on each token’s modality. Furthermore, DAMEX~\cite{Jain2023DAMEXDM} introduced a dataset-aware MoE model where experts specialize in specific datasets rather than modalities, learning to route input data based on its dataset of origin.

\paragraph{Remote Sensing Foundation Models}

Remote sensing enables Earth observation through active and passive sensors mounted on satellites or aerial platforms, capturing vast amounts of heterogeneous data~\cite{lillesand2015remote, toth2016remote}. Unlike traditional RGB imagery, satellite data varies significantly across sensor types, making model transferability challenging. To address this, foundation models for remote sensing leverage large-scale self-supervised pretraining on unlabeled data. Most foundation models target specific modalities, such as optical RGB~\cite{Cong2022SatMAEPT, Reed2022ScaleMAEAS}, multispectral~\cite{hong2024spectralgpt}, hyperspectral~\cite{wang2024hypersigma}, or SAR~\cite{glaser2024wv} images. More recent works, such as MMEarth~\cite{Nedungadi2024MMEarthEM}, DOFA~\cite{Xiong2024NeuralPM}, RingMo-SAM~\cite{yan2023ringmo}, SkySense~\cite{guo2024skysense}, and SSL-based multi-modal models~\cite{scheibenreif2022self}, explore large-scale multi-modal pretraining by integrating diverse sensor data to improve generalization. However, these models typically remain monolithic, processing all available modalities. In contrast, MAPEX introduces a selective Mixture-of-Experts approach that activates only the relevant experts per input, enabling the extraction of smaller, task-specific models for downstream applications. This improves computational efficiency while preserving strong multi-modal performance.

\section{Methods}
Figure \ref{fig:overview} presents a high-level overview of our method.
\vspace{-1em}
\paragraph{Preliminaries} We consider a multi-modal dataset $\mathcal{D}_n^m$ consisting of $n$ co-located and temporally matched samples from $m$ different modalities. Our model is based on the vision transformer (ViT) base architecture~\cite{dosovitskiy2020image}. Input samples $\mathbf{x}_i^j\sim\mathcal{D}_n^m$ of shape $\mathbf{x}\in\mathbb{R}^{(\sum_j^m C^j)\times H\times W}$ are split into patches and linearly embedded per modality with learnable patch embedding transforms $\{f^j\}_{j=1,...,m}$. This yields a set of $m$ $D$-dimensional tokens for each sample patch. Sinusoidal positional embeddings to encode spatial information are added to each token~\cite{vaswani2017attention}. We investigate sinusoidal and learnable \textit{modality}-tokens to encode the modality of tokens across the channel dimension. To mark modality transitions in the token sequence, learnable end-of-modality tokens are injected before attention operations.

\subsection{Masked Image Modeling}
\label{sec:mae}
% \begin{itemize}
%     \item modality-wise patch embed
%     \item modality-wise masking
%     \item modality-wise decoding
%     \item modality dropout
% \end{itemize}

We utilize masked image modeling as self-supervised pre-training technique~\cite{he2022masked}. After patch-embedding, a binary mask is applied to the input sequence of tokens, independently masking $75\%$ of the tokens corresponding to each modality. The remaining tokens are processed by a ViT encoder. Once encoded, the original sequence length is restored by introducing special, learnable, \texttt{[MASK]} tokens at the positions of dropped input tokens. A ViT decoder then aims to reconstruct the original (normalized) pixel values of the masked patches. We use a mean-squared error objective between pixel values of the original masked patches and their reconstruction for training~\cite{he2022masked}.
\paragraph{Modality-wise Decoding} To keep the reconstruction task difficult in the presence of multiple independently masked modalities (a spatial patch might be masked in one modality but visible in another), we reconstruct all modalities independently. All tokens of a modality are processed by the decoder in a distinct forward pass, incurring a total of $m$ decoder forward passes, while also reducing the sequence length in each pass by a factor of $\frac{1}{m}$.

\subsection{Mixture-of-Modality Experts}
\label{sec:moe}
% \begin{itemize}
%     \item modality routing
%     \item deterministic routing
%     \item topk routing
%     \item learned modality routing
%     \item load-balancing loss
%     \item universal expert
%     \item modality dropout
% \end{itemize}
Mixture-of-experts (MoE) models replace the feed-forward layers in each ViT block with a routing module and multiple feed-forward \textit{expert} modules~\cite{shazeer2017outrageously}. The router consists of a linear layer that produces logits for each token, from which a softmax distribution over the $e$ experts is constructed. The expert outputs for each token are then weighted by the routing softmax weights. We adapt the MoE framework to enforce $\text{expert}{\xleftrightarrow{}}\text{modality}$ relationships in the model. Starting from a fixed assignment of experts to modalities, we ultimately design a learnable routing mechanism based on modality information that consistently routes tokens from a modality $M$ to the same subset of experts. Top-$k$ selection of the $k$ highest ranked experts is used to achieve sparse conditional computation for each modality.
Our \textbf{deterministic routing} setup defines a fixed one-to-one relationship between modalities and experts. Every modality is assigned to exactly one expert a priori, and no experts are shared between modalities.
\textbf{Positional embedding routing} utilizes sinusoidal positional embeddings~\cite{vaswani2017attention} across the stack of modalities as input to the routing module. While the routing weights are learned, the input embeddings remain fixed and unique for each modality. This approach supports an arbitrary number of experts but introduces a potentially inappropriate ordering between modalities.
\textbf{Modality routing} introduces learnable [\texttt{MODALITY}] embeddings that represent each modality in lieu of channel-wise positional embeddings. These tokens are added to each modality and used as input to the routing module. Explicitly, for token $t^m$ of modality $m$, the routing weights are computed by passing the associated modality embedding to the routing module $r$: $w_r^m = r($[\texttt{MODALITY}]$^m)$. Unlike standard MoE implementations, this routing strategy guarantees that tokens of a modality will always be routed to the same experts, even across samples.
\paragraph{Load Balancing Loss} Imbalanced expert utilization is a common problem with learned routing mechanisms~\cite{shazeer2017outrageously}. We observe that some experts randomly receive a lower shares of tokens at the beginning of training, resulting in fewer gradient updates and continued low utilization of these experts throughout training. We address this issue by introducing a differentiable load-balancing loss that encourages a more uniform distribution of routing probabilities between experts:
\begin{equation}
    \mathcal{L}_{\text{load}}=\frac{1}{e}\sum_{i=1}^e(U_i - \frac{1}{e})^2,
\end{equation}
with $e$ the total number of experts and $U_i$ the empirical utilization fraction of the $i^\text{th}$ expert (i.e., the fraction of all tokens that expert $i$ processes). This additional loss term encourages more uniform expert utilization.
During pre-training, the load balancing loss is combined with the reconstruction loss to create the training objective: $\mathcal{L} = \mathcal{L_\text{rec}} + \alpha\cdot\mathcal{L_\text{load}}$ where the weighting factor $\alpha$ scales the load balancing loss.
\paragraph{Shared Expert} We hypothesize that some features are shared between modalities and introduce an additional shared expert that always processes tokens from all modalities. This allows the modality experts to focus on the specific information for each modality, while common features like edges can be detected by the shared expert.
\paragraph{Modality Dropout} In order to prime the model for downstream applications where the full set of pre-training modalities is not available, we utilize modality dropout during the pre-training stage. In each batch, the tokens corresponding to a pre-defined fraction of modalities are set to zero.

\subsection{Downstream Expert Pruning}
% \begin{itemize}
%    \item pruning based on routing probabilities
%    \item ...?
% end{itemize}
After pre-training the mixture-of-modality experts model with masked autoencoding on unlabeled multi-modal data, we transfer the model encoder to the downstream task of interest. Often only a subset of the pre-training modalities will be available for the downstream task. Accordingly, we prune the mixture-of-modality experts to suit the available data modalities. For a model pre-trained with experts $E^p$ on $\mathcal{M}^p$ pre-training modalities and $k$ experts per modality, all experts and patch-embedding projections corresponding to modalities that are not in the set of downstream modalities are dropped. The downstream experts $E^d$ are extracted from each pre-trained model layer as:
\begin{equation}
    E^d = \{E_i | i \in\argtopk(w_r^m) \forall m \in \mathcal{M}^d\}_{i=1,...,|E^p|},
\end{equation}
with $w_r^m$ the pre-training routing probabilities for modality $m$. In the case of a single downstream modality, this yields a model with $k$ downstream experts.

% \begin{figure}
%     \centering
%     \includegraphics[width=1.\linewidth]{figures/moe_routing.pdf}
%     \caption{Modality embedding based routing. Routing probabilities for each embedded patch are computed by the router based on the embedding of the patch's modality. Expert outputs are aggregated over experts weighted by their routing probabilities.}
%     \label{fig:enter-label}
% \end{figure}

\begin{table*}[t]
    \centering
    % \scalebox{0.85}{
    \begin{tabular}{l|c|c|c|c|c|c|c}
        \toprule
         \multirow{2}{*}{Modality} & \multirow{2}{*}{Supervised FS} & \multicolumn{2}{c|}{SatMAE \cite{Cong2022SatMAEPT}} & \multicolumn{2}{c|}{Scale-MAE \cite{Reed2022ScaleMAEAS}} &  \multicolumn{2}{c}{{MAPEX} (Ours)} \\
         % \cline{3-8}
         &  & $k$-NN & FT & $k$-NN & FT & $k$-NN & FT  \\
         \midrule
         Architecture & \textit{MAPEX} & \multicolumn{2}{c|}{\textit{vit\_base} } & \multicolumn{2}{c|}{\textit{vit\_large}} & \multicolumn{2}{c}{\textit{MAPEX}} \\
         \# params & 130M & \multicolumn{2}{c|}{90M} & \multicolumn{2}{c|}{310M} & \multicolumn{2}{c}{130M} \\
         \midrule
         RGB &  71.2 & 64.5 & 70.8 & 75.1 & \textbf{79.9} & 67.2 & 75.6 \\
         Red Edge  & 73.3 & 65.5 & 69.7 & 74.7 & \textbf{80.3} & 66.7 & 77.8\\
         SWIR  & 76.5 & 68.3 & 74.5 & 76.2 & \textbf{83.7} & 72.4 & 79.8 \\
         NIR  & 56.1 & 50.0 & 56.9 & 49.9 & 57.2 & 56.8 & \textbf{59.4}\\
         SAR  & 69.7 & 62.2 & 67.2 & 65.3 & 73.1 & 66.1 & \textbf{73.6}\\
         ELE  & 51.8 & 46.5 & 47.1 & 46.6 & 51.4 & 48.2 & \textbf{58.8}\\
         \bottomrule
    \end{tabular}%}
    \caption{$k$-NN and fine-tuning (FT) performance (accuracy in \%) for land-cover classification on different modalities of the \textit{ben-ge-8k} dataset~\cite{mommert2023ben}. For each modality, the best performing model is highlighted in \textbf{bold}.}
    \label{tab:downstream_sota}
\end{table*}

\section{Datasets}
% \begin{itemize}
%     \item benge
%     \item benge-8k land cover
%     \item SEN12-floods
%     \item wildfire
%     \item ...?
% \end{itemize}

\paragraph{BigEarthNet-GeoEnvironment 
(ben-ge)} The \textit{ben-ge} dataset~\cite{mommert2023ben} is an extension to the BigEarthNet-MM~\cite{sumbul2021bigearthnet} dataset. It consists of $590\,326$ colocated and temporally matched samples from Sentinel-1 (SAR), Sentinel-2 (multi-spectral) as well as elevation and environmental data. We utilize the \textit{ben-ge} dataset for multi-modal pre-training of the mixture-of-modality experts model. We group the data into six distinct modalities: \textit{RGB} (Sentinel-2 B04, B03, B02), \textit{Red Edge} (Sentinel-2 B05, B06, B07), \textit{SWIR} (short-wave infrared, Sentinel-2 B11, B12, B8A), \textit{NIR} (near-infrared, Sentinel-2 B08), \textit{SAR} (Sentinel-1 VV and VH polarizations), and \textit{ELE} (elevation map). \textbf{ben-ge-8k} is a subset of \textit{ben-ge} consisting of ${\approx} 8\,000$ samples that we use for downstream evaluation on multi-label land-use/land-cover classification with different modalities.
\paragraph{SEN12-FLOOD} This dataset consists of Sentinel-1 and Sentinel-2 imagery acquired in areas prone to flooding and corresponding binary flood labels~\cite{rambour2020sen12} across the world. It contains 366 time-series of varying length, with $512{\times}512$ pixels per tile. The dataset is split by locations and tiles within a time-series are considered independently. We use the SAR data from SEN12-FLOOD for downstream evaluation of the pre-trained model on a single modality.
\vspace{-1em}
\paragraph{California Wildfire}
We construct a dataset specifically for wildfire detection by collecting fire perimeter data from the California Fire Perimeters dataset~ \cite{california_fire_perimeters} and recent CAL FIRE incidents records~ \cite{cal_fire_incidents}. These records include detailed spatial and temporal information on wildfire events, such as locations, dates, and sizes. For each fire perimeter, we acquire corresponding Sentinel-2 satellite imagery~\cite{drusch2012sentinel}. To create a binary classification dataset, we crop small image patches ($224{\times}224$ pixels) and label them as either \texttt{wildfire} or \texttt{no-wildfire} based on overlap with the wildfire perimeter. We utilize Sentinel-2 short-wave infrared (SWIR) bands (B8A, B11, and B12), which have proven most effective in detecting burned areas \cite{nbr, cocke2005comparison}. The resulting dataset contains $20\,000$ samples, with $11\,000$ labeled as \texttt{wildfire} and $9\,000$ as \texttt{no-wildfire}.

\section{Experiments and Results}

\subsection{Downstream Evaluation}
\label{sec:experiments}
% Establish that this approach produces good models for fine-tuning on specific modalities. I.e., as good as other methods for the modalities they are trained on, better on all other modalities.
\paragraph{Baselines}
We evaluate pre-trained {MAPEX} across several downstream tasks and compare to other remote sensing foundation models. {MAPEX} is based on the ViT-base architecture and retains two experts after pruning for one modality. SatMAE~\cite{Cong2022SatMAEPT} is a ViT-base model trained on multispectral Sentinel-2 data, while Scale-MAE~\cite{Reed2022ScaleMAEAS} is trained on the RGB bands of Sentinel-2. Since Scale-MAE only provides pre-trained weights for a ViT-large architecture, we use that configuration in our comparisons. Our setup is as follows: after pre-training, we retain only the top two FFN experts for each modality, pruning the others. This results in smaller, modality-specialized models (\ie RGB experts for RGB input, SAR experts for SAR input, \etc). These specialized models are  also compared against the same model architecture and size (2 experts) trained fully supervised from scratch (FS).
% We report both $k$-Nearest Neighbors ($k$-NN) classification and finetuning (FT) accuracy to assess different training level.
\vspace{-1em}
\paragraph{Specialized Models} Table \ref{tab:downstream_sota} shows that fine-tuning our specialized models on their respective input modalities outperforms other methods for NIR, SAR, and ELE inputs. This pattern is consistent whether we perform $k$-NN classification on features extracted from the specialized models or fine-tune the entire models.  On modalities that are closer to Scale-MAE pre-training data, this model outperforms our approach. This is partly explained by the fact that Scale-MAE is $2.3{\times}$ larger than our specialized models. More importantly, our method outperforms the bigger Scale-MAE on those modalities that differ the most from the pre-training data used for Scale-MAE.

\begin{table}[t]
    \centering
    \scalebox{0.82}{
    \begin{tabular}{l|c|c|c|c|c|c|c}
        \toprule
         \multirow{2}{*}{Modality} & \multirow{2}{*}{FS} & \multicolumn{2}{c|}{SatMAE} &
         \multicolumn{2}{c|}{Scale-MAE} & \multicolumn{2}{c}{{MAPEX}} \\
         % \cline{2-7}
         &  & $k$-NN & FT & $k$-NN & FT  & $k$-NN & FT \\
         \midrule
         Architecture & \textit{MAPEX} & \multicolumn{2}{c|}{\textit{vit\_base}} & \multicolumn{2}{c|}{\textit{vit\_large}} & \multicolumn{2}{c}{\textit{MAPEX}} \\
         \# params & 130M & \multicolumn{2}{c|}{90M} & \multicolumn{2}{c|}{310M} & \multicolumn{2}{c}{130M} \\
         \midrule
         SAR & 78.1 & 77.4 & 80.3 & 83.0 & 82.6 & 80.1 & \textbf{83.2} \\
         \bottomrule
    \end{tabular}}
    \caption{$k$-NN and fine-tuning performance (accuracy in \%) for binary flood detection using Sentinel 1 data~\cite{rambour2020sen12}.}
    \label{tab:flood_detection}
\end{table}

\begin{wrapfigure}{r}{0.3\textwidth}
    \centering
    \includegraphics[width=1\linewidth]{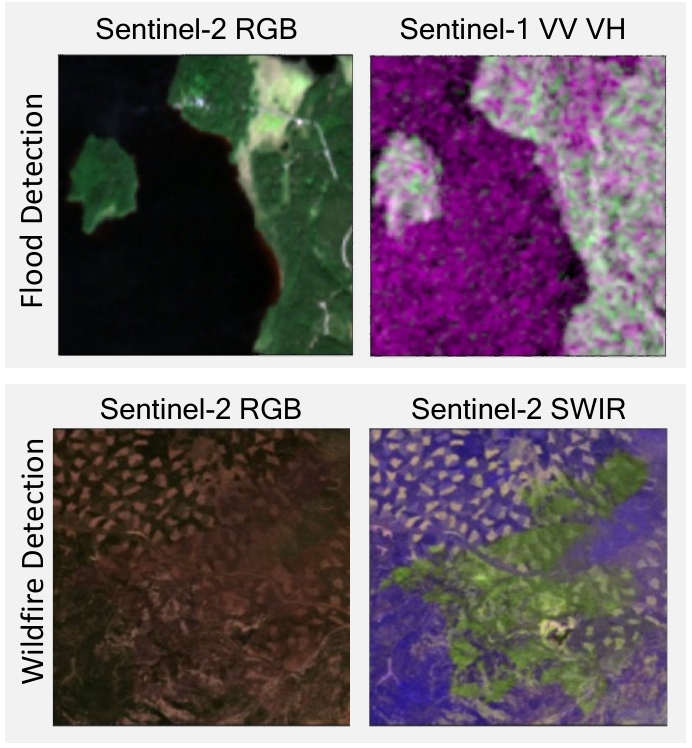}
    \caption{
    Task specific modalities: Sentinel-1 bands provide valuable information for detecting water bodies in the presence of clouds. Similarly, SWIR bands are more effective than RGB for identifying burned areas.}
    \label{fig:data-samples}
    \vspace{-1em}
\end{wrapfigure}

\begin{table}
    \centering

    \scalebox{0.82}{
    \begin{tabular}{l|c|c|c|c|c|c|c}
        \toprule
         \multirow{2}{*}{Modality} & \multirow{2}{*}{FS} & \multicolumn{2}{c|}{SatMAE} &
          \multicolumn{2}{c|}{Scale-MAE} & \multicolumn{2}{c}{MAPEX} \\
         % \cline{2-7}
         &  & $k$-NN & FT &  $k$-NN & FT & $k$-NN & FT  \\
         \midrule
         Architecture & \textit{MAPEX} & \multicolumn{2}{c|}{\textit{vit\_base}} & \multicolumn{2}{c|}{\textit{vit\_large}} & \multicolumn{2}{c}{\textit{MAPEX}} \\
         \# params & 130M & \multicolumn{2}{c|}{90M} & \multicolumn{2}{c|}{310M} & \multicolumn{2}{c}{130M} \\
         \midrule
         SWIR  & 84.7 & 84.2 & 88.6 & 86.6 & \textbf{90.8} & 87.3 & 90.5 \\
         \bottomrule
    \end{tabular}}
     \caption{$k$-NN and fine-tuning performance for Californian wildfire detection \cite{california_fire_perimeters} using short-wave infrared (SWIR) data (Sentinel-2 B8A, B11, B12 bands).}
    \label{tab:wildfire}
\end{table}
\begin{table}
    \centering
    \scalebox{0.9}{
    \begin{tabular}{lccc}
        \toprule
         Modality & U-Net \cite{Ronneberger2015UNetCN}  & SatMAE (\textit{vit\_base}) & MAPEX \\
         \midrule
         RGB & 56.3 & 53.1 & \textbf{56.9}  \\
         \bottomrule
    \end{tabular}}
     \caption{mIoU (in \%) for semantic segmentation on the \textit{ben-ge-8k} dataset \cite{mommert2023ben}}
    \label{tab:segmentation}
    % \vspace{-1.4em}
\end{table}
Besides land-cover classification, we test {MAPEX} on two other specific tasks: flood and wildfire detection. These are important applications that are commonly tackled with specific remote sensing modalities (see Figure~\ref{fig:data-samples}).
%selected to illustrate how certain data modalities contains more relevant information for specific tasks (see Figure~ \ref{fig:data-samples}).

\begin{table}[b]
    \centering

    \scalebox{0.92}{
    \begin{tabular}{lccc}
        \toprule
         Modality &  Dropout = 0 &  Dropout = 0.1 &  Dropout = 0.5 \\
         \midrule
         RGB & 65.8 & 65.8 & \textbf{67.2}  \\
         Red Edge & 66.1 & 66.2 & \textbf{66.7}  \\
         SWIR & 71.6 & 71.9 & \textbf{72.4}  \\
         NIR & 54.2 & 54.3 & \textbf{56.8}  \\
         SAR & 65.7 & 65.8 & \textbf{66.1}  \\
         ELE & 46.6 & 47.1 & \textbf{48.2}  \\
         \bottomrule
    \end{tabular}}
    \caption{Comparison of different modality dropout values. $k$-NN accuracy for land-cover classification on different modalities of the ben-ge-8k dataset~\cite{mommert2023ben} is reported.}
    \label{tab:dropouts}
    \vspace{-1em}
\end{table}

For example, SAR (Sentinel-1) data is more informative than RGB data for detecting water in cloudy weather, while the SWIR bands of Sentinel-2 provide better information than RGB data for detecting burned areas. Since most foundation models are commonly trained on either the full spectrum of Sentinel-2 or only on RGB data, we demonstrate in Tables~\ref{tab:flood_detection} and~\ref{tab:wildfire} that our smaller, specialized models outperform general foundation models. Specifically, for flood detection, {MAPEX} outperforms SatMAE by 3\% (absolute) and ScaleMAE by 1\%, while for wildfire detection, it surpasses SatMAE by 2\% and performs comparably to ScaleMAE despite being considerably smaller. Beyond classification and detection, we also evaluate MAPEX on segmentation tasks to assess its ability to handle dense predictions. To this end, we add the same simple segmentation head to both the RGB-specific pruned model (with MAPEX) and SatMAE, and we compare the results to a U-Net baseline. Table \ref{tab:segmentation} shows that MAPEX achieves the best performance, demonstrating its versatility across multiple remote sensing tasks.
\vspace{-1em}
\paragraph{Shared Expert } We study the impact of using a shared expert on MAPEX performance across multiple modalities. The shared expert is hypothesized to capture common cross-modal information, enabling modality-specific experts to specialize in unique features. Table \ref{tab:univ_experts} indicates that, while the shared expert generally enhances performance across most modalities, it also introduces a significant increase in model parameters. This raises a trade-off between performance improvement and computational efficiency, suggesting that the shared expert may not always provide a cost-effective benefit, particularly in resource-constrained applications.

\begin{table}[b]
    \centering

    % \scalebox{0.88}{
    \begin{tabular}{lcc}
        \toprule
         Modality & 2 Experts &  2 Experts + 1 shared \\
         \midrule
         \# params & 130M & 200M \\
         \midrule
         RGB & 67.2 & \textbf{67.6}   \\
         Red Edge & 66.7 & \textbf{67.1}   \\
         SWIR & 72.4 & \textbf{72.9}   \\
         NIR & \textbf{56.8} & 56.0   \\
         SAR & 66.1 & \textbf{66.7}   \\
         ELE & \textbf{48.2} &  47.8  \\
         \bottomrule
    \end{tabular}%}
    \caption{$k$-NN performance for land-cover classification on different modalities of the ben-ge-8k dataset~\cite{mommert2023ben}, with and without the use of a shared expert.}
    \label{tab:univ_experts}
\end{table}

\begin{figure}[t]
    \centering
    \includegraphics[width=0.9
\linewidth]{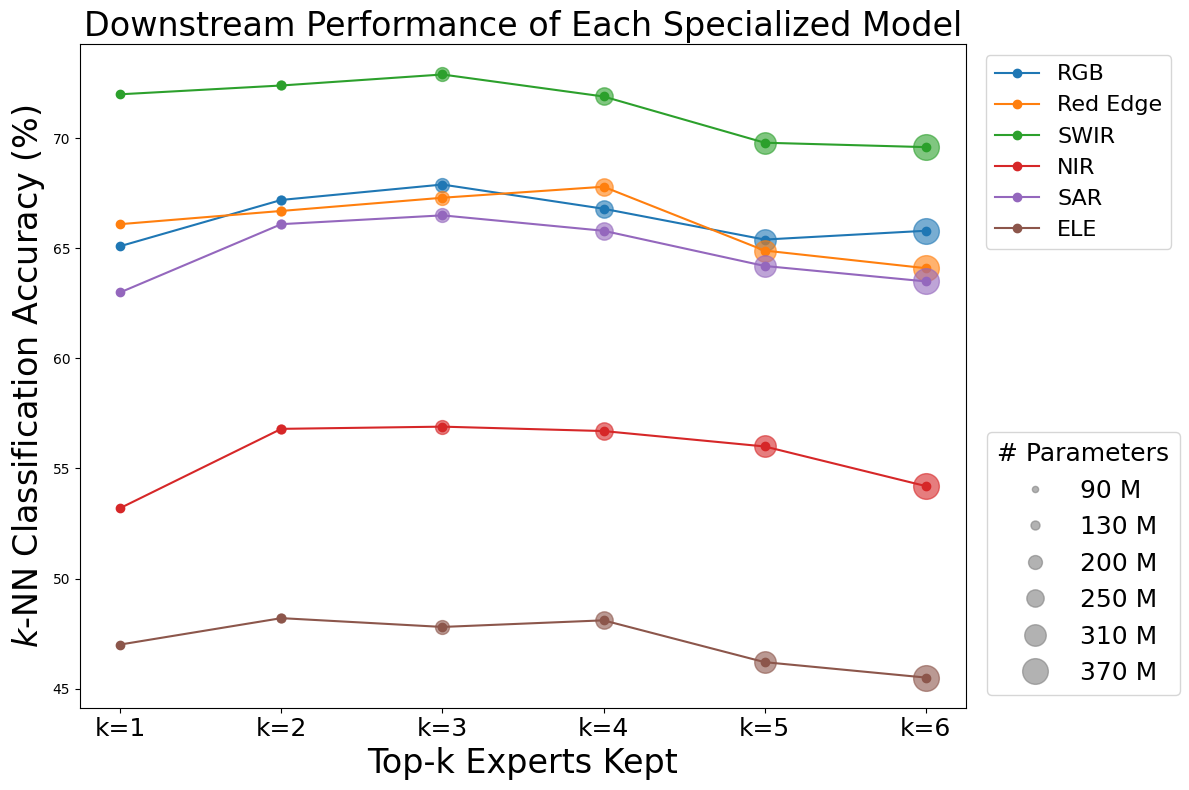}
    \caption{Downstream accuracy with different numbers of retained modality experts $k$.
    % Effect of top-$k$ experts on model performance across modalities.
    Most models achieve peak accuracy with top-$k{\in}\{2,3\}$, while more experts tend to reduce accuracy.}
    \label{fig:topk}
     \vspace{-1.3em}
\end{figure}

\begin{figure}[b]
    \centering
    \includegraphics[width=0.85\linewidth]{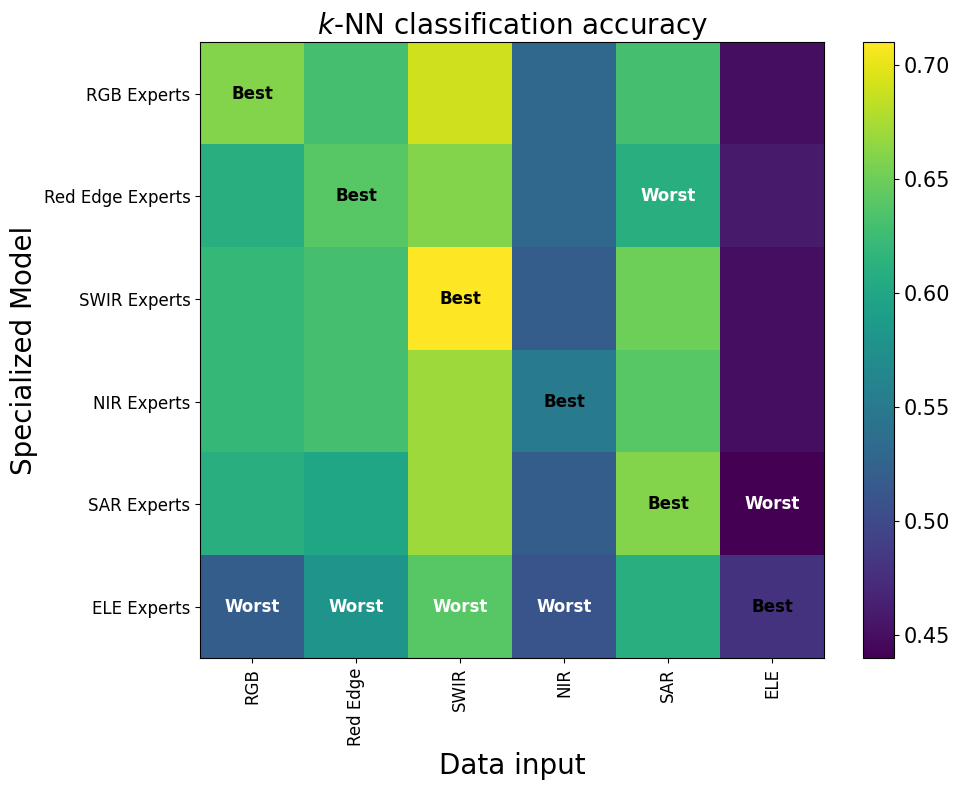}
    \caption{$\text{Expert}{\xleftrightarrow{}}\text{modality}$ specialization on the \textit{ben-ge-8k} dataset~\cite{mommert2023ben}. Each row corresponds to the performance of {MAPEX} pruned for a specific modality and applied across different modalities. The best and worst specialized models for each input modality are highlighted.}
    \label{fig:knn_acc_experts}
    \vspace{-1.2em}
\end{figure}

\paragraph{Modality Dropout } We investigate the impact of varying modality dropout values (0\%, 10\%, and 50\%) on downstream model performance. Table \ref{tab:dropouts} shows that a dropout value of 50\% outperforms both 10\% and 0\% by around 1.5\% (absolute) on average. Intuitively, a higher dropout value encourages the model to become more robust by learning representations that do not rely on the presence of all modalities. By setting a modality dropout value of 50\% during pre-training, the model learns to handle partial data input, which aligns better with scenarios where some modalities may be unavailable. This approach is particularly beneficial as a significant portion of the model is shared during pre-training, with only certain parts pruned for downstream tasks. As a result, the shared component gains robustness from the dropout, enhancing the model's adaptability across different downstream specialized tasks.
\vspace{-1em}
\paragraph{Number of Experts } We explore how the number of selected experts (top-$k$) influences the downstream performance of specialized models across the different modalities. Figure \ref{fig:topk} shows that most modality-specific models achieve their highest accuracy when using the top 2 or 3 experts. Higher top-$k$ values often reduce accuracy likely because the model becomes too large, with too much capacity for a small dataset. Given that, we choose the top 2 as it yields a smaller, more efficient model with minimal loss in accuracy.
\vspace{-1.5em}
\paragraph{Specialization of Experts} This work focuses on transforming a general, modality-agnostic foundation model into smaller modality-specific models by selectively keeping experts specialized to each type of data. To test the effectiveness of these specialized models, we create modality-specific versions and evaluate them across all available data types. Results are shown in Figure~\ref{fig:knn_acc_experts}. This involves, for example, retaining only the RGB experts to build an RGB-specific model, then testing it with various input modalities (represented in the first row of Figure~\ref{fig:knn_acc_experts}). We repeat this process for each modality. The heatmap reveals that each specialized model performs best when tested on its intended data type (e.g., RGB experts on RGB data, NIR experts on NIR data). These results indicate that {MAPEX} expert routing successfully instills $\text{expert}{\xleftrightarrow{}}\text{modality}$  relationships into the model during pre-training.
%while a model trained on RGB data can handle other data types to some extent, it cannot match the accuracy of a model specialized for each specific modality.

\begin{figure}[t]
    \centering
    \includegraphics[width=1.0
\linewidth]{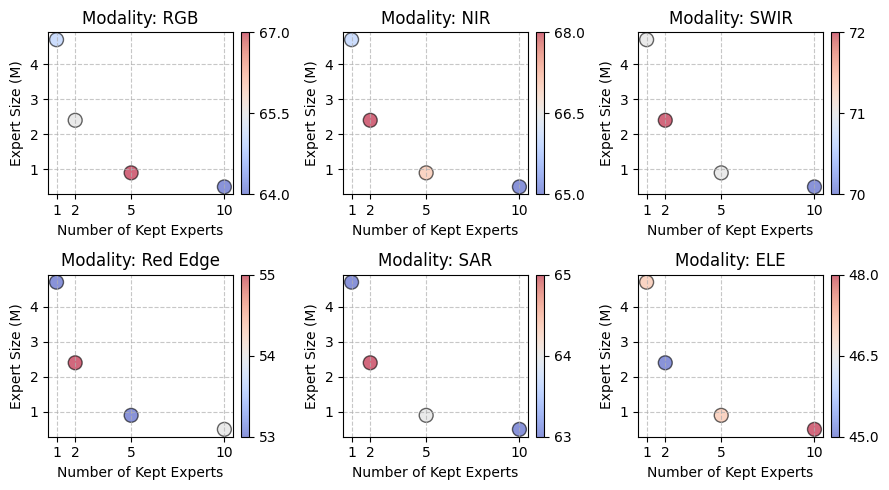}
    \caption{Downstream accuracy (\%) for different expert numbers and sizes per modality. Total model size is fixed at 90M.}
    \label{fig:expert-size}
     \vspace{-1.3em}
\end{figure}

\paragraph{Balancing Expert Size and Number} To find the ideal number of experts and sizes, we conduct an experiment with fixed downstream model size (90M, ViT-Base) while varying the number and size of experts per modality. Models are pretrained with different numbers of experts and pruned to a fixed number per modality: 1, 2, 5, or 10 experts are retained per modality. This corresponds to expert sizes of 4.7M, 2.4M, 0.9M, and 0.5M parameters, respectively. Figure \ref{fig:expert-size} shows that optimal expert configurations vary by modality, but the majority of modalities achieve the highest accuracy with 2 experts per modality. For RGB, accuracy peaks with 5 smaller experts (0.9M each), suggesting that moderate specialization enhances learning. In contrast, elevation (ELE) performs best with fewer, larger experts, indicating a less diverse feature space that benefits from stronger individual experts. These findings suggest that a fixed number of experts per modality is suboptimal, but 2 experts per modality is a good compromise across most cases.

\paragraph{Implementation Details}
% \begin{itemize}
%     \item classification head
%     \item k-nn approach
%     \item optimizer, lr, etc.
%     \item number of epochs
%     \item image size, augmentations
% \end{itemize}

To evaluate a pre-trained model with $k$-NN classification, we extract and average-pool image-level features for each sample. Then, a $k$-NN classifier with $k{=}5$ is fit to the feature representations. For fine-tuning on classification tasks, we append an additional linear classification layer to the pre-trained model.
For pre-training and downstream evaluation, images are resized to $224{\times}224$ pixels and normalized channel-wise. We fine-tune the models for 30 epochs, using AdamW \cite{Loshchilov2017DecoupledWD} optimizer and a learning rate scheduler that reduces the rate when the validation loss ceases to improve.

\begin{table}[t]
    \centering

    \scalebox{0.92}{
    \begin{tabular}{lccc}
        \toprule
         Modality & Deterministic & Pos. Embed & Modality-based \\
         \midrule
         RGB & 63.5 & 63.1 & \textbf{67.2}  \\
         Red Edge & 63.2 & 64.9 & \textbf{66.7}  \\
         SWIR & 68.7 & 67.9 & \textbf{72.4}  \\
         NIR & 51.6 & 55.9 & \textbf{56.8}  \\
         SAR & 62.2 & 65.3 & \textbf{66.1}  \\
         ELE & 46.0 & \textbf{48.3} & 48.2  \\
         \bottomrule
    \end{tabular}}
     \caption{Comparison of routing mechanisms using $k$-NN accuracy for land-cover classification on different \textit{ben-ge-8k}~\cite{mommert2023ben} modalities.}
    \label{tab:routings}
    \vspace{-1.2em}
\end{table}

% \begin{table}[]
%     \centering
%     \caption{Comparison of different topk values}
%     \scalebox{0.88}{
%     \begin{tabular}{lcccccc}
%         \toprule
%          Modality & topk=1 & topk=2 & topk=3 & topk=4 & topk=5 &  topk=6 \\
%          \midrule
%          \# params & 90M & 130M & 200M & 250M & 310M & 370M \\
%          \midrule
%          RGB & xx & 67.2 & xx & xx & xx & xx \\
%          Red Edge & xx & 66.7 & xx & xx & xx & xx \\
%          SWIR & xx & 72.4 & xx & xx & xx & xx \\
%          NIR & xx & 56.8 & xx & xx & xx & xx \\
%          SAR & xx & 66.1 & xx & xx & xx & xx \\
%          ELE & xx & 48.2 & xx & xx & xx & xx \\
%          \bottomrule
%     \end{tabular}}
%     \label{tab:topk}
% \end{table}

\subsection{Self-supervised Pre-training}
% Show that this approach of pre-training and pruning creates modality-specific models. Pruning does "destroy" the model.
% \begin{itemize}
%     \item Show reconstructions?
%     \item compare with uni-modal pretraining
%     \item Motivate need for aux loss
    % \item Show specialization of experts (heatmap plot, i.e. we pretrain + prune in the right way)
% \end{itemize}

\begin{figure}
    \centering
    \includegraphics[width=1\linewidth]{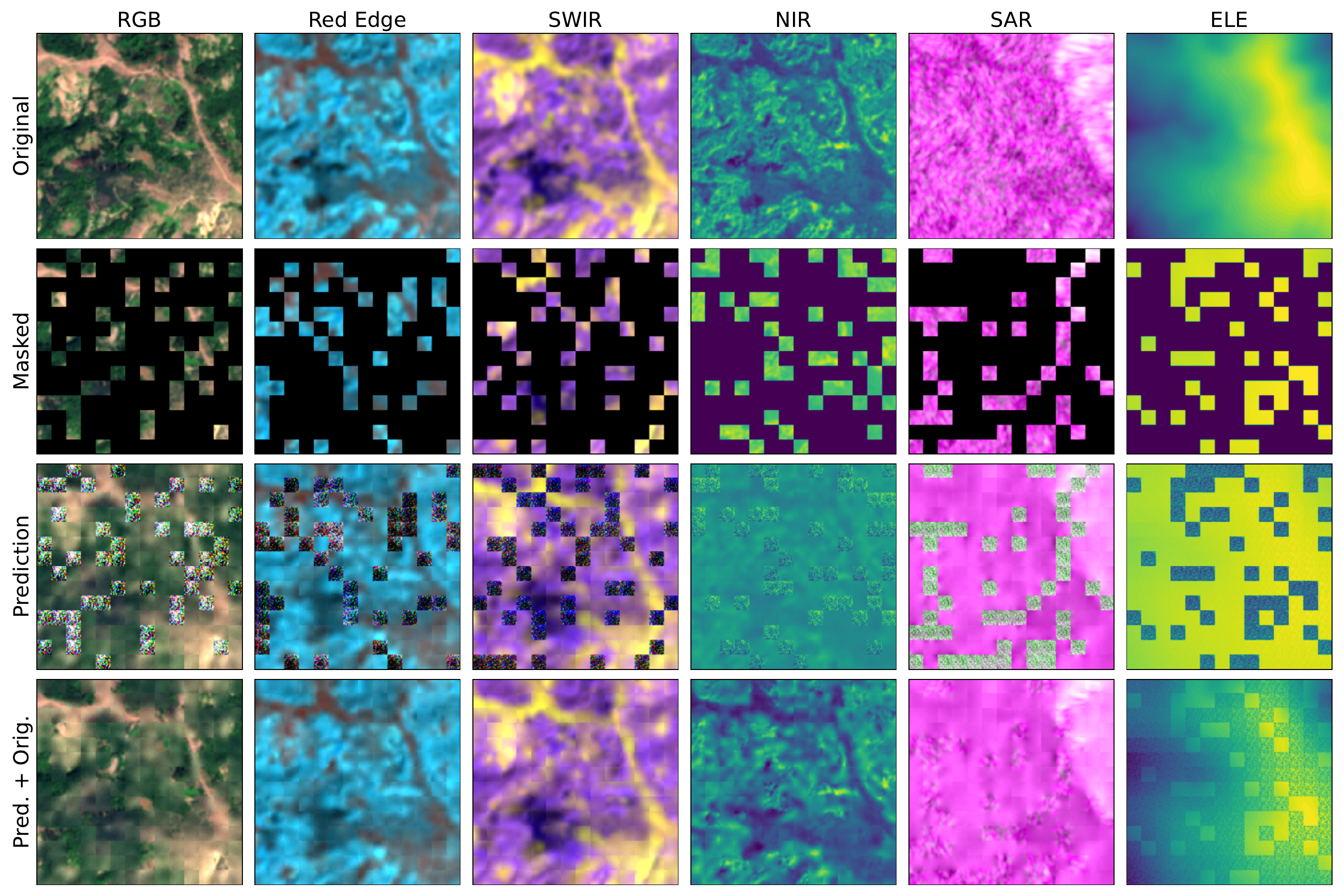}
    \caption{Reconstruction examples from mixture-of-modality experts model on multi-modal data. From left to right: optical RGB, red edge, short-wave infrared, near infrared, synthetic aperture radar, and elevation from \textit{ben-ge}~\cite{mommert2023ben}. The same decoder is used for all modalities.}
    \label{fig:mm_recon}
    \vspace{-1em}
\end{figure}

{MAPEX} models are pre-trained with masked autoencoding on multi-modal remote sensing data. Representations generated by the mixture-of-modality experts encoder (6 experts, $360\text{M}$ parameters) are mapped back to pixel-space for each modality with an 8-block transformer decoder (see Section~\ref{sec:mae}). We pre-train for 50 epochs with AdamW~\cite{Loshchilov2017DecoupledWD} on the \textit{ben-ge} dataset. The load balancing loss is weighted by $\alpha{=}0.01$ which yields near uniform expert utilization.
Pre-training results in good reconstruction performance (see Fig.~\ref{fig:mm_recon}) for most modalities. Reconstruction is most difficult for the SAR backscatter data, where mean-squared reconstruction error remains ${\approx}3\times$ higher than on the other modalities.

% \begin{figure*}[t]
%     \centering
%     \includegraphics[width=0.8\linewidth]{figures/mm_reconstruction2.pdf}
%     \caption{Reconstruction examples from mixture-of-modality experts model on multi-modal data. From left to right: optical RGB, red edge, short-wave infrared, near infrared, synthetic aperture radar, and elevation from \textit{ben-ge}~\cite{mommert2023ben}. The same decoder is used for all modalities.}
%     \label{fig:mm_recon}
%     \vspace{-1em}
% \end{figure*}

\vspace{-1em}
\paragraph{Routing Mechanism}
In contrast to standard MoE approaches, this work aims to elicit strong $\text{expert}\xleftrightarrow{}\text{modality}$ relationships during pre-training, which facilitates expert-pruning for downstream tasks. In standard token-based routing, routing probabilities for each token $t$ are computed by the router as $w_r{=}r(t)$. Initial experiments with token-based routing on multi-modal remote sensing data failed to produce a clear specialization of experts for individual modalities, and suffered from unbalanced expert utilization. We address and evaluate these shortcomings with three alternative routing mechanisms (see Section~\ref{sec:moe}): deterministic routing, positional embedding routing, and modality-based routing.
We find that pre-training with modality-based routing consistently outperforms the two other routing mechanisms across downstream tasks of different modalities (see Table~\ref{tab:routings}). This confirms our hypothesis that the routing parameters should be learnable (unlike the deterministic routing), and that learnable modality tokens are a suitable input for the routing module.

\paragraph{Multi-modal Pre-training} We assess the effect of multi-modal pre-training on the \textit{ben-ge} dataset (see Table~\ref{tab:pretrain_strategy}). Downstream performance on the RGB modality improves with multi-modal pre-training compared to pre-training only on the downstream modality.
\vspace{-1em}

\begin{wraptable}{r}{0.25\textwidth}
\scalebox{0.8}{
    \begin{tabular}{c|c|c}
    \toprule
    Modalities & Experts & $k\text{-NN}$ Acc.\\
    \midrule
    1 & 1 & $62.1$\\
    6 & 1 & $63.2$\\
    6 & 6 & $65.1$\\
    \bottomrule
    \end{tabular}
    }
    \caption{Effect of uni vs. multi-modal pre-training and number of pre-training experts on \textit{ben-ge-8k}. Accuracy on the RGB modality after pruning all models to 1 expert.}
    \label{tab:pretrain_strategy}
    \vspace{-1em}
\end{wraptable}

\paragraph{Expert Modules} To evaluate the benefit of using multiple expert modules in multi-modal data, we compare multi-modal {MAPEX} variants with a single and multiple experts (see Table~\ref{tab:pretrain_strategy}). After pre-training, both models are pruned to one expert and evaluated on the RGB modality. The model pre-trained with multiple experts significantly outperforms the model with one expert. Please note that both models are the same size after the pruning stage.

%\noindent
%These results first validate the utility of pre-training on multimodal data. Even though the model is evaluated on RGB input alone, the shared layers are likely enriched with cross-modal representations, which improves the performance. Second, the results highlight the importance of starting with a model of higher capacity, allowing it to specialize through training, and then retaining only the relevant components.

% \subsection{Multi-modal Downstream Task}
% Combine RGB+S1 or something like that?

% \subsection{Maybe Few-shot FT?}

\section{Discussion \& Conclusion}
% \begin{itemize}
%     \item feasibility of multi-modal pre-training
%     \item specialization of experts
%     \item performance after pruning
%     \item downstream performance / handling
%     \item broader impact and limitations
% \end{itemize}

This work proposes a mixture-of-modality experts module suitable for large-scale self-supervised pre-training on multi-modal data. To improve downstream performance on individual modalities, we present a pruning mechanism to derive efficient downstream models specialized for individual modalities from the pre-trained model.
The {MAPEX} model is successfully pre-trained on multi-modal remote sensing data, achieving good reconstruction performance across modalities. Our experiments show that the modality-based expert routing mechanism elicits specialization among expert modules, resulting in sub-networks that can be mapped to individual modalities.
% \begin{wraptable}{r}{0.25\textwidth}
% \scalebox{0.8}{
%     \begin{tabular}{c|c|c}
%     \toprule
%     Modalities & Experts & $k\text{-NN}$ Acc.\\
%     \midrule
%     1 & 1 & $62.1$\\
%     6 & 1 & $63.2$\\
%     6 & 6 & $65.1$\\
%     \bottomrule
%     \end{tabular}
%     }
%     \caption{Effect of uni vs. multi-modal pre-training and number of pre-training experts on \textit{ben-ge-8k}. Accuracy on the RGB modality after pruning all models to 1 expert.}
%     \label{tab:pretrain_strategy}
%     \vspace{-1em}
% \end{wraptable}
When pruned for a specific modality, the large-scale pre-trained model turns into a much smaller specialized model that performs well on downstream tasks without any fine-tuning, as $k$-NN evaluation experiments show. Fine-tuned, our modality-specific models perform competitively with existing fine-tuned remote sensing foundation models.
%\vspace{-2em}

\noindent\textbf{Broader Impact and Opportunities} Foundation models are transforming the way remote sensing data is processed. However, large models are costly to fine-tune and deploy. Additionally, remote sensing tasks are often label-scarce, or utilize modalities where pre-trained foundation models do not exist yet. Our work addresses these use-cases of remote sensing foundation models by incorporating many modalities during pre-training, and provides ways to remove model parts trained for expendable or unavailable modalities for downstream applications.
{MAPEX} offers straightforward extension to additional modalities, but remains limited by the availability of data from the target modality during pre-training.

{
    \small
    \bibliographystyle{ieeenat_fullname}
    \bibliography{main}

\begin{thebibliography}{51}
\providecommand{\natexlab}[1]{#1}
\providecommand{\url}[1]{\texttt{#1}}
\expandafter\ifx\csname urlstyle\endcsname\relax
  \providecommand{\doi}[1]{doi: #1}\else
  \providecommand{\doi}{doi: \begingroup \urlstyle{rm}\Url}\fi

\bibitem[Amitrano et~al.(2024)Amitrano, Di~Martino, Di~Simone, and Imperatore]{amitrano2024flood}
Donato Amitrano, Gerardo Di~Martino, Alessio Di~Simone, and Pasquale Imperatore.
\newblock {Flood Detection with SAR: A review of Techniques and Datasets}.
\newblock \emph{Remote Sensing}, 16\penalty0 (4):\penalty0 656, 2024.

\bibitem[Baltru{\v{s}}aitis et~al.(2018)Baltru{\v{s}}aitis, Ahuja, and Morency]{baltruvsaitis2018multimodal}
Tadas Baltru{\v{s}}aitis, Chaitanya Ahuja, and Louis-Philippe Morency.
\newblock {Multimodal Machine Learning: A Survey and Taxonomy}.
\newblock \emph{IEEE Transactions on Pattern Analysis and Machine Intelligence}, 41\penalty0 (2):\penalty0 423--443, 2018.

\bibitem[Bommasani et~al.(2021)Bommasani, Hudson, Adeli, Altman, Arora, von Arx, Bernstein, Bohg, Bosselut, Brunskill, et~al.]{bommasani2021opportunities}
Rishi Bommasani, Drew~A Hudson, Ehsan Adeli, Russ Altman, Simran Arora, Sydney von Arx, Michael~S Bernstein, Jeannette Bohg, Antoine Bosselut, Emma Brunskill, et~al.
\newblock {On the Opportunities and Risks of Foundation Models}.
\newblock \emph{arXiv preprint arXiv:2108.07258}, 2021.

\bibitem[Brown(2020)]{brown2020language}
Tom~B Brown.
\newblock {Language Models are Few-shot Learners}.
\newblock \emph{arXiv preprint arXiv:2005.14165}, 2020.

\bibitem[{California Department of Forestry and Fire Protection}({\natexlab{a}})]{cal_fire_incidents}
{California Department of Forestry and Fire Protection}.
\newblock {CAL FIRE Incidents}.
\newblock \url{https://www.fire.ca.gov/incidents}, {\natexlab{a}}.
\newblock Accessed: 2024-11.

\bibitem[{California Department of Forestry and Fire Protection}({\natexlab{b}})]{california_fire_perimeters}
{California Department of Forestry and Fire Protection}.
\newblock {California Fire Perimeters (all)}.
\newblock \url{https://catalog.data.gov/dataset/california-fire-perimeters-all-b3436}, {\natexlab{b}}.
\newblock Accessed: 2024-11.

\bibitem[Cao et~al.(2024)Cao, You, and Liu]{cao2024mv}
Jingyi Cao, Yanan You, and Jun Liu.
\newblock {MV-MOE: A Visual Mixture-of-Experts Model for Optical-SAR Image Matching}.
\newblock In \emph{IGARSS 2024-2024 IEEE International Geoscience and Remote Sensing Symposium}, pages 9676--9679. IEEE, 2024.

\bibitem[Chi et~al.(2016)Chi, Plaza, Benediktsson, Sun, Shen, and Zhu]{chi2016big}
Mingmin Chi, Antonio Plaza, Jon~Atli Benediktsson, Zhongyi Sun, Jinsheng Shen, and Yangyong Zhu.
\newblock {Big Data for Remote Sensing: Challenges and Opportunities}.
\newblock \emph{Proceedings of the IEEE}, 104\penalty0 (11):\penalty0 2207--2219, 2016.

\bibitem[Cocke et~al.(2005)Cocke, Ful{\'e}, and Crouse]{cocke2005comparison}
Allison~E Cocke, Peter~Z Ful{\'e}, and Joseph~E Crouse.
\newblock {Comparison of Burn Severity Assessments using Differenced Normalized Burn Ratio and Ground Data}.
\newblock \emph{International Journal of Wildland Fire}, 14\penalty0 (2):\penalty0 189--198, 2005.

\bibitem[Cong et~al.(2022)Cong, Khanna, Meng, Liu, Rozi, He, Burke, Lobell, and Ermon]{Cong2022SatMAEPT}
Yezhen Cong, Samarth Khanna, Chenlin Meng, Patrick Liu, Erik Rozi, Yutong He, M. Burke, D. Lobell, and Stefano Ermon.
\newblock {SatMAE: Pre-training Transformers for Temporal and Multi-Spectral Satellite Imagery}.
\newblock \emph{ArXiv}, abs/2207.08051, 2022.

\bibitem[Dosovitskiy(2020)]{dosovitskiy2020image}
Alexey Dosovitskiy.
\newblock {An Image is Worth 16x16 Words: Transformers for Image Recognition at Scale}.
\newblock \emph{arXiv preprint arXiv:2010.11929}, 2020.

\bibitem[Drusch et~al.(2012)Drusch, Del~Bello, Carlier, Colin, Fernandez, Gascon, Hoersch, Isola, Laberinti, Martimort, et~al.]{drusch2012sentinel}
Matthias Drusch, Umberto Del~Bello, S{\'e}bastien Carlier, Olivier Colin, Veronica Fernandez, Ferran Gascon, Bianca Hoersch, Claudia Isola, Paolo Laberinti, Philippe Martimort, et~al.
\newblock {Sentinel-2: ESA's optical high-resolution mission for GMES operational services}.
\newblock \emph{Remote sensing of Environment}, 120:\penalty0 25--36, 2012.

\bibitem[Fedus et~al.(2022)Fedus, Zoph, and Shazeer]{fedus2022switch}
William Fedus, Barret Zoph, and Noam Shazeer.
\newblock {Switch Transformers: Scaling to Trillion Parameter Models with Simple and Efficient Sparsity}.
\newblock \emph{Journal of Machine Learning Research}, 23\penalty0 (120):\penalty0 1--39, 2022.

\bibitem[{Geospatial Data Platform}()]{nbr}
{Geospatial Data Platform}.
\newblock {Normalized burn ratio (NBR)}.
\newblock \url{https://docs.up42.com/help/spectral-indexes/nbr}.
\newblock Accessed: 2024-11.

\bibitem[Glaser et~al.(2024)Glaser, Stopa, Wolniewicz, Foster, Vandemark, Mouche, Chapron, and Sadowski]{glaser2024wv}
Yannik Glaser, Justin~E Stopa, Linnea~M Wolniewicz, Ralph Foster, Doug Vandemark, Alexis Mouche, Bertrand Chapron, and Peter Sadowski.
\newblock {WV-Net: A Foundation Model for SAR WV-mode Satellite Imagery Trained using Contrastive Self-supervised Learning on 10 Million Images}.
\newblock \emph{arXiv preprint arXiv:2406.18765}, 2024.

\bibitem[Guo et~al.(2024)Guo, Lao, Dang, Zhang, Yu, Ru, Zhong, Huang, Wu, Hu, et~al.]{guo2024skysense}
Xin Guo, Jiangwei Lao, Bo Dang, Yingying Zhang, Lei Yu, Lixiang Ru, Liheng Zhong, Ziyuan Huang, Kang Wu, Dingxiang Hu, et~al.
\newblock {Skysense: A Multi-modal Remote Sensing Foundation Model Towards Universal Interpretation for Earth Observation Imagery}.
\newblock In \emph{Proceedings of the IEEE/CVF Conference on Computer Vision and Pattern Recognition}, pages 27672--27683, 2024.

\bibitem[Han et~al.()Han, Zhang, Shi, and Reichstein]{hanmultisensory}
Boran Han, Shuai Zhang, Xingjian Shi, and Markus Reichstein.
\newblock {Multisensory Geospatial Models via Cross-Sensor Pretraining}.

\bibitem[Hanna et~al.(2023)Hanna, Mommert, and Borth]{Hanna2023SparseMV}
Joelle Hanna, Michael Mommert, and Damian Borth.
\newblock Sparse multimodal vision transformer for weakly supervised semantic segmentation.
\newblock \emph{2023 IEEE/CVF Conference on Computer Vision and Pattern Recognition Workshops (CVPRW)}, pages 2145--2154, 2023.

\bibitem[He et~al.(2022)He, Chen, Xie, Li, Doll{\'a}r, and Girshick]{he2022masked}
Kaiming He, Xinlei Chen, Saining Xie, Yanghao Li, Piotr Doll{\'a}r, and Ross Girshick.
\newblock {Masked Autoencoders are Scalable Vision Learners}.
\newblock In \emph{Proceedings of the IEEE/CVF Conference on Computer Vision and Pattern Recognition}, pages 16000--16009, 2022.

\bibitem[Hinton(2015)]{hinton2015distilling}
Geoffrey Hinton.
\newblock {Distilling the Knowledge in a Neural Network}.
\newblock \emph{arXiv preprint arXiv:1503.02531}, 2015.

\bibitem[Hong et~al.(2024)Hong, Zhang, Li, Li, Li, Yao, Yokoya, Li, Ghamisi, Jia, et~al.]{hong2024spectralgpt}
Danfeng Hong, Bing Zhang, Xuyang Li, Yuxuan Li, Chenyu Li, Jing Yao, Naoto Yokoya, Hao Li, Pedram Ghamisi, Xiuping Jia, et~al.
\newblock {SpectralGPT: Spectral Remote Sensing Foundation Model}.
\newblock \emph{IEEE Transactions on Pattern Analysis and Machine Intelligence}, 2024.

\bibitem[Hu et~al.(2021)Hu, Shen, Wallis, Allen-Zhu, Li, Wang, Wang, and Chen]{hu2021lora}
Edward~J Hu, Yelong Shen, Phillip Wallis, Zeyuan Allen-Zhu, Yuanzhi Li, Shean Wang, Lu Wang, and Weizhu Chen.
\newblock {LoRA: Low-rank Adaptation of Large Language Models}.
\newblock \emph{arXiv preprint arXiv:2106.09685}, 2021.

\bibitem[Jain et~al.(2023)Jain, Behl, Kira, and Vineet]{Jain2023DAMEXDM}
Yash Jain, Harkirat~Singh Behl, Zsolt Kira, and Vibhav Vineet.
\newblock Damex: Dataset-aware mixture-of-experts for visual understanding of mixture-of-datasets.
\newblock \emph{ArXiv}, abs/2311.04894, 2023.

\bibitem[Kaplan et~al.(2020)Kaplan, McCandlish, Henighan, Brown, Chess, Child, Gray, Radford, Wu, and Amodei]{kaplan2020scaling}
Jared Kaplan, Sam McCandlish, Tom Henighan, Tom~B Brown, Benjamin Chess, Rewon Child, Scott Gray, Alec Radford, Jeffrey Wu, and Dario Amodei.
\newblock {Scaling Laws for Neural Language Models}.
\newblock \emph{arXiv preprint arXiv:2001.08361}, 2020.

\bibitem[Kirillov et~al.(2023)Kirillov, Mintun, Ravi, Mao, Rolland, Gustafson, Xiao, Whitehead, Berg, Lo, et~al.]{kirillov2023segment}
Alexander Kirillov, Eric Mintun, Nikhila Ravi, Hanzi Mao, Chloe Rolland, Laura Gustafson, Tete Xiao, Spencer Whitehead, Alexander~C Berg, Wan-Yen Lo, et~al.
\newblock {Segment Anything}.
\newblock In \emph{Proceedings of the IEEE/CVF International Conference on Computer Vision}, pages 4015--4026, 2023.

\bibitem[Li et~al.(2024)Li, Gan, Yang, Yang, Li, Wang, Gao, et~al.]{li2024multimodal}
Chunyuan Li, Zhe Gan, Zhengyuan Yang, Jianwei Yang, Linjie Li, Lijuan Wang, Jianfeng Gao, et~al.
\newblock {Multimodal Foundation Models: From Specialists to General-purpose Assistants}.
\newblock \emph{Foundations and Trends{\textregistered} in Computer Graphics and Vision}, 16\penalty0 (1-2):\penalty0 1--214, 2024.

\bibitem[Lillesand et~al.(2015)Lillesand, Kiefer, and Chipman]{lillesand2015remote}
Thomas Lillesand, Ralph~W Kiefer, and Jonathan Chipman.
\newblock \emph{{Remote Sensing and Image Interpretation}}.
\newblock John Wiley \& Sons, 2015.

\bibitem[Lin et~al.(2024{\natexlab{a}})Lin, Hong, Ge, Luo, Jiang, Jin, and Wen]{lin2024rs}
Hui Lin, Danfeng Hong, Shuhang Ge, Chuyao Luo, Kai Jiang, Hao Jin, and Congcong Wen.
\newblock {RS-MoE: Mixture of Experts for Remote Sensing Image Captioning and Visual Question Answering}.
\newblock \emph{arXiv preprint arXiv:2411.01595}, 2024{\natexlab{a}}.

\bibitem[Lin et~al.(2024{\natexlab{b}})Lin, Shrivastava, Luo, Iyer, Lewis, Gosh, Zettlemoyer, and Aghajanyan]{lin2024moma}
Xi~Victoria Lin, Akshat Shrivastava, Liang Luo, Srinivasan Iyer, Mike Lewis, Gargi Gosh, Luke Zettlemoyer, and Armen Aghajanyan.
\newblock {MoMa: Efficient Early-fusion Pre-training with Mixture of Modality-aware Experts}.
\newblock \emph{arXiv preprint arXiv:2407.21770}, 2024{\natexlab{b}}.

\bibitem[Loshchilov and Hutter(2017)]{Loshchilov2017DecoupledWD}
Ilya Loshchilov and Frank Hutter.
\newblock {Decoupled Weight Decay Regularization}.
\newblock In \emph{International Conference on Learning Representations}, 2017.

\bibitem[Lu et~al.(2024)Lu, Guo, Zimmer-Dauphinee, Nieusma, Wang, VanValkenburgh, Wernke, and Huo]{lu2024ai}
Siqi Lu, Junlin Guo, James~R Zimmer-Dauphinee, Jordan~M Nieusma, Xiao Wang, Parker VanValkenburgh, Steven~A Wernke, and Yuankai Huo.
\newblock {AI Foundation Models in Remote Sensing: A Survey}.
\newblock \emph{arXiv preprint arXiv:2408.03464}, 2024.

\bibitem[Mommert et~al.(2023)Mommert, Kesseli, Hanna, Scheibenreif, Borth, and Demir]{mommert2023ben}
Michael Mommert, Nicolas Kesseli, Jo{\"e}lle Hanna, Linus Scheibenreif, Damian Borth, and Beg{\"u}m Demir.
\newblock {Ben-ge: Extending BigEarthNet with geographical and environmental data}.
\newblock In \emph{IGARSS 2023-2023 IEEE International Geoscience and Remote Sensing Symposium}, pages 1016--1019. IEEE, 2023.

\bibitem[Nedungadi et~al.(2024{\natexlab{a}})Nedungadi, Kariryaa, Oehmcke, Belongie, Igel, and Lang]{nedungadi2024mmearth}
Vishal Nedungadi, Ankit Kariryaa, Stefan Oehmcke, Serge Belongie, Christian Igel, and Nico Lang.
\newblock {MMEarth: Exploring Multi-Modal Pretext Tasks for Geospatial Representation Learning}.
\newblock \emph{arXiv preprint arXiv:2405.02771}, 2024{\natexlab{a}}.

\bibitem[Nedungadi et~al.(2024{\natexlab{b}})Nedungadi, Kariryaa, Oehmcke, Belongie, Igel, and Lang]{Nedungadi2024MMEarthEM}
Vishal Nedungadi, Ankit Kariryaa, Stefan Oehmcke, Serge~J. Belongie, Christian Igel, and Nico Lang.
\newblock Mmearth: Exploring multi-modal pretext tasks for geospatial representation learning.
\newblock \emph{ArXiv}, abs/2405.02771, 2024{\natexlab{b}}.

\bibitem[Radford et~al.(2021)Radford, Kim, Hallacy, Ramesh, Goh, Agarwal, Sastry, Askell, Mishkin, Clark, et~al.]{radford2021learning}
Alec Radford, Jong~Wook Kim, Chris Hallacy, Aditya Ramesh, Gabriel Goh, Sandhini Agarwal, Girish Sastry, Amanda Askell, Pamela Mishkin, Jack Clark, et~al.
\newblock {Learning Transferable Visual Models from Natural Language Supervision}.
\newblock In \emph{International Conference on Machine Learning}, pages 8748--8763. PMLR, 2021.

\bibitem[Rambour et~al.(2020)Rambour, Audebert, Koeniguer, Le~Saux, Crucianu, and Datcu]{rambour2020sen12}
C Rambour, N Audebert, E Koeniguer, B Le~Saux, M Crucianu, and M Datcu.
\newblock {Sen12-flood: a SAR and Multispectral Dataset for Flood Detection}.
\newblock \emph{IEEE: Piscataway, NJ, USA}, 2020.

\bibitem[Reed et~al.(2022)Reed, Gupta, Li, Brockman, Funk, Clipp, Candido, Uyttendaele, and Darrell]{Reed2022ScaleMAEAS}
Colorado Reed, Ritwik Gupta, Shufan Li, Sara Brockman, Christopher Funk, Brian Clipp, Salvatore Candido, Matthew Uyttendaele, and Trevor Darrell.
\newblock {Scale-MAE: A Scale-Aware Masked Autoencoder for Multiscale Geospatial Representation Learning}.
\newblock \emph{2023 IEEE/CVF International Conference on Computer Vision (ICCV)}, pages 4065--4076, 2022.

\bibitem[Riquelme et~al.(2021)Riquelme, Puigcerver, Mustafa, Neumann, Jenatton, Susano~Pinto, Keysers, and Houlsby]{riquelme2021scaling}
Carlos Riquelme, Joan Puigcerver, Basil Mustafa, Maxim Neumann, Rodolphe Jenatton, Andr{\'e} Susano~Pinto, Daniel Keysers, and Neil Houlsby.
\newblock {Scaling Vision with Sparse Mixture of Experts}.
\newblock \emph{Advances in Neural Information Processing Systems}, 34:\penalty0 8583--8595, 2021.

\bibitem[Ronneberger et~al.(2015)Ronneberger, Fischer, and Brox]{Ronneberger2015UNetCN}
Olaf Ronneberger, Philipp Fischer, and Thomas Brox.
\newblock U-net: Convolutional networks for biomedical image segmentation.
\newblock \emph{ArXiv}, abs/1505.04597, 2015.

\bibitem[Scheibenreif et~al.(2022)Scheibenreif, Hanna, Mommert, and Borth]{scheibenreif2022self}
Linus Scheibenreif, Jo{\"e}lle Hanna, Michael Mommert, and Damian Borth.
\newblock {Self-supervised Vision Transformers for Land-cover Segmentation and Classification}.
\newblock In \emph{Proceedings of the IEEE/CVF Conference on Computer Vision and Pattern Recognition}, pages 1422--1431, 2022.

\bibitem[Shazeer et~al.(2017)Shazeer, Mirhoseini, Maziarz, Davis, Le, Hinton, and Dean]{shazeer2017outrageously}
Noam Shazeer, Azalia Mirhoseini, Krzysztof Maziarz, Andy Davis, Quoc Le, Geoffrey Hinton, and Jeff Dean.
\newblock {Outrageously Large Neural Networks: The Sparsely-gated Mixture-of-Experts Layer}.
\newblock \emph{arXiv preprint arXiv:1701.06538}, 2017.

\bibitem[Sietsma and Dow(1988)]{sietsma1988neural}
Sietsma and Dow.
\newblock {Neural Net Pruning-Why and How}.
\newblock In \emph{IEEE 1988 international conference on neural networks}, pages 325--333. IEEE, 1988.

\bibitem[Sishodia et~al.(2020)Sishodia, Ray, and Singh]{sishodia2020applications}
Rajendra~P Sishodia, Ram~L Ray, and Sudhir~K Singh.
\newblock {Applications of Remote Sensing in Precision Agriculture: A Review}.
\newblock \emph{Remote sensing}, 12\penalty0 (19):\penalty0 3136, 2020.

\bibitem[Sumbul et~al.(2021)Sumbul, De~Wall, Kreuziger, Marcelino, Costa, Benevides, Caetano, Demir, and Markl]{sumbul2021bigearthnet}
Gencer Sumbul, Arne De~Wall, Tristan Kreuziger, Filipe Marcelino, Hugo Costa, Pedro Benevides, Mario Caetano, Beg{\"u}m Demir, and Volker Markl.
\newblock {BigEarthNet-MM: A large-scale, multimodal, multilabel benchmark archive for remote sensing image classification and retrieval}.
\newblock \emph{IEEE Geoscience and Remote Sensing Magazine}, 9\penalty0 (3):\penalty0 174--180, 2021.

\bibitem[Toth and J{\'o}{\'z}k{\'o}w(2016)]{toth2016remote}
Charles Toth and Grzegorz J{\'o}{\'z}k{\'o}w.
\newblock {Remote Sensing Platforms and Sensors: A Survey}.
\newblock \emph{ISPRS Journal of Photogrammetry and Remote Sensing}, 115:\penalty0 22--36, 2016.

\bibitem[Vaswani(2017)]{vaswani2017attention}
A Vaswani.
\newblock {Attention is all you need}.
\newblock \emph{Advances in Neural Information Processing Systems}, 2017.

\bibitem[Wang et~al.(2024)Wang, Hu, Jin, Miao, Yang, Xu, Qin, Ma, Sun, Li, et~al.]{wang2024hypersigma}
Di Wang, Meiqi Hu, Yao Jin, Yuchun Miao, Jiaqi Yang, Yichu Xu, Xiaolei Qin, Jiaqi Ma, Lingyu Sun, Chenxing Li, et~al.
\newblock {HyperSIGMA: Hyperspectral Intelligence Comprehension Foundation Model}.
\newblock \emph{arXiv preprint arXiv:2406.11519}, 2024.

\bibitem[Xiao et~al.(2024)Xiao, Xuan, Wang, Huang, Tao, Lu, and Yokoya]{xiao2024foundation}
Aoran Xiao, Weihao Xuan, Junjue Wang, Jiaxing Huang, Dacheng Tao, Shijian Lu, and Naoto Yokoya.
\newblock {Foundation Models for Remote Sensing and Earth Observation: A Survey}.
\newblock \emph{arXiv preprint arXiv:2410.16602}, 2024.

\bibitem[Xiong et~al.(2024)Xiong, Wang, Zhang, Stewart, Hanna, Borth, Papoutsis, Saux, Camps-Valls, and Zhu]{Xiong2024NeuralPM}
Zhitong Xiong, Yi Wang, Fahong Zhang, Adam~J. Stewart, Joelle Hanna, Damian Borth, Ioannis Papoutsis, Bertrand~Le Saux, Gustau Camps-Valls, and Xiao~Xiang Zhu.
\newblock Neural plasticity-inspired multimodal foundation model for earth observation.
\newblock 2024.

\bibitem[Yan et~al.(2023)Yan, Li, Li, Zhou, Zhang, Feng, Diao, Fu, and Sun]{yan2023ringmo}
Zhiyuan Yan, Junxi Li, Xuexue Li, Ruixue Zhou, Wenkai Zhang, Yingchao Feng, Wenhui Diao, Kun Fu, and Xian Sun.
\newblock {RingMo-SAM: A Foundation Model for Segment Anything in Multimodal Remote-sensing Images}.
\newblock \emph{IEEE Transactions on Geoscience and Remote Sensing}, 61:\penalty0 1--16, 2023.

\bibitem[Zhu et~al.(2017)Zhu, Tuia, Mou, Xia, Zhang, Xu, and Fraundorfer]{zhu2017deep}
Xiao~Xiang Zhu, Devis Tuia, Lichao Mou, Gui-Song Xia, Liangpei Zhang, Feng Xu, and Friedrich Fraundorfer.
\newblock {Deep Learning in Remote Sensing: A Comprehensive Review and List of Resources}.
\newblock \emph{IEEE Geoscience and Remote Sensing Magazine}, 5\penalty0 (4):\penalty0 8--36, 2017.

\end{thebibliography}
}

\clearpage
\setcounter{page}{1}
\maketitlesupplementary

% \section{Rationale}
% \label{sec:rationale}
% %
% Having the supplementary compiled together with the main paper means that:
% %
% \begin{itemize}
% \item The supplementary can back-reference sections of the main paper, for example, we can refer to \cref{sec:intro};
% \item The main paper can forward reference sub-sections within the supplementary explicitly (e.g. referring to a particular experiment);
% \item When submitted to arXiv, the supplementary will already included at the end of the paper.
% \end{itemize}
% %
% To split the supplementary pages from the main paper, you can use \href{https://support.apple.com/en-ca/guide/preview/prvw11793/mac#:~:text=Delete%20a%20page%20from%20a,or%20choose%20Edit%20%3E%20Delete).}{Preview (on macOS)}, \href{https://www.adobe.com/acrobat/how-to/delete-pages-from-pdf.html#:~:text=Choose%20%E2%80%9CTools%E2%80%9D%20%3E%20%E2%80%9COrganize,or%20pages%20from%20the%20file.}{Adobe Acrobat} (on all OSs), as well as \href{https://superuser.com/questions/517986/is-it-possible-to-delete-some-pages-of-a-pdf-document}{command line tools}.

\begin{strip}
% \begin{table*}[!ht]
    \centering
    \begin{tabular}{l|c|c|c|c}
        \toprule
         \multirow{2}{*}{} & Supervised FS & SatMAE \cite{Cong2022SatMAEPT} & ScaleMAE \cite{Reed2022ScaleMAEAS} &  MAPEX (Ours) \\
         % \cline{3-8}
         \midrule
         Architecture & \textit{MAPEX} & \textit{vit\_base} & \textit{vit\_large} & \textit{MAPEX} \\
         \# params & 130M & 90M & 310M & 130M \\
         \midrule
         10-shot & 32.1 & 34.5 & 38.9 & 38.6   \\
         100-shot  & 48.1 & 53.0 & 57.9 & 56.1  \\
         \bottomrule
    \end{tabular}
    \captionof{table}{Few-shot results for fine-tuning (accuracy in \%) on land-cover classification using the RGB modality of \textit{ben-ge-8k} \cite{mommert2023ben}.}
    % \caption
    \label{tab:few_shot_supp}
% \end{table*}
\begin{tabular}{l|c|c|c|c|c|c|c}
        \toprule
         \multirow{2}{*}{Modality} & \multirow{2}{*}{Supervised FS} & \multicolumn{2}{c|}{SatMAE \cite{Cong2022SatMAEPT}} & \multicolumn{2}{c|}{ScaleMAE \cite{Reed2022ScaleMAEAS}} &  \multicolumn{2}{c}{{MAPEX} (Ours)} \\
         % \cline{3-8}
         &  & $k$-NN & FT & $k$-NN & FT & $k$-NN & FT  \\
         \midrule
         Architecture & \textit{MAPEX} & \multicolumn{2}{c|}{\textit{vit\_base} } & \multicolumn{2}{c|}{\textit{vit\_large}} & \multicolumn{2}{c}{\textit{MAPEX}} \\
         \# params & 130M & \multicolumn{2}{c|}{90M} & \multicolumn{2}{c|}{310M} & \multicolumn{2}{c}{130M} \\
         \midrule
         RGB &  71.2 & 64.5 & 70.8 & 75.1 & 79.9 & 67.2 & 75.6 \\
         SAR  & 69.7 & 62.2 & 67.2 & 65.3 & 73.1 & 66.1 & 73.6\\
         RGB + SAR  & 74.8 & 70.8 & 74.7 & 67.2 & 80.2 & 72.9 & 79.9\\
         \bottomrule
    \end{tabular}
    \captionof{table}{$k$-NN and fine-tuning (FT) performance (accuracy in \%) for land-cover classification on different modalities of the \textit{ben-ge-8k} dataset~\cite{mommert2023ben}. For each modality, the best performing model is highlighted in \textbf{bold}.}
    \label{tab:more_modalitites_supp}
\end{strip}

\section{Few-shot Learning}
\label{sec:sup:few_shot}

We perform few-shot experiments to evaluate the label efficiency of our method and compare its performance with other approaches. In each k-shot experiment, we randomly select k samples for every class from the training set. We keep the same experimental setup as described in Section \ref{sec:experiments}. Table~\ref{tab:few_shot_supp} reports the performance of MAPEX compared to SatMAE \cite{Cong2022SatMAEPT} and ScaleMAE \cite{Reed2022ScaleMAEAS} on the RGB modality of the \textit{ben-ge-8k} dataset. The results, presented for both 10-shot and 100-shot scenarios, show that MAPEX achieves competitive performance, outperforming SatMAE in both scenarios and nearing the performance of ScaleMAE, despite ScaleMAE having a significantly larger architecture. This demonstrates that MAPEX can handle limited labeled data effectively, making it a practical and efficient alternative to larger models like ScaleMAE.

\section{Multimodal Downstream Evaluation}
\label{sec:sup:multimodal}
Since combining multiple modalities is often beneficial for some downstream tasks, we evaluate the multimodal capabilities of MAPEX by testing its performance on different data modalities: RGB, SAR, and their combination (RGB + SAR) from the \textit{ben-ge-8k} dataset. For unimodal inputs, we use the top-2 experts per modality, while for the combined RGB + SAR input, we use the top-1 expert per modality to maintain the same model size during evaluation. Note that for the multimodal setup, we merge the data channel-wise before feeding it to the model. Table \ref{tab:more_modalitites_supp} shows that MAPEX outperforms SatMAE and the supervised baseline for both RGB and SAR inputs while closely matching ScaleMAE, despite its significantly smaller architecture. These results show that MAPEX can easily adapt and compete with other methods, depending on the available modalities at test time.

\end{document}